# Leveraging cross-platform data to improve automated hate speech detection

John D. Gallacher*
Oxford Internet Institute
University of Oxford

## Abstract

Hate speech is increasingly prevalent online, and its negative outcomes include increased prejudice, extremism, and even offline hate crime. Automatic detection of online hate speech can help us to better understand these impacts. However, while the field has recently progressed through advances in natural language processing, challenges still remain. In particular, most existing approaches for hate speech detection focus on a single social media platform in isolation. This limits both the use of these models and their validity, as the nature of language varies from platform to platform. Here we propose a new cross-platform approach to detect hate speech which leverages multiple datasets and classification models from different platforms and trains a 'superlearner' that can combine existing and novel training data to improve detection and increase model applicability. We demonstrate how this approach outperforms existing models, and achieves good performance when tested on messages from novel social media platforms not included in the original training data.

* **Corresponding Author:** John Gallacher – john.gallacher@oii.ox.ac.uk

# Introduction

While the Internet has allowed people from across the globe to connect and communicate, it has also given those who wish to spread abuse, incivility, hate, and other forms of extreme digital speech the unprecedented ability to do so, and at greater scale than ever. Up to 1% of content on mainstream social media platforms now contains some form of extreme digital speech, and as many as 40% of online users report having seen some form of extreme online speech (Vidgen, Margetts, & Harris, 2020), suggesting that it receives disproportionate attention. This reach is increasing, with online hate crimes rising by 40% since 2017 (Home Office, 2018; Williams & Mishcon de Reya, 2019).

Given this increasing prevalence, there is a growing requirement to understand both the nature and impact of online hate speech. Analysis of online abuse and hate speech requires robust methods to detect and measure it, and manual qualitative analysis of individual messages cannot scale to the billions of posts made online on social media every day (Internet Live Stats, 2020), limiting its application. In addition, exposing human moderators to the harmful nature of extreme digital speech can be damaging to mental health over the long term (Gillespie, 2018; King, 2018; Roberts, 2019). Because of this, automated techniques to detect extreme digital speech are essential for both research and practical applications.

Automatically detecting extreme digital speech in online communications faces a number of challenges, however (Vidgen, Tromble, et al., 2019). These include how to accurately detect harmful content across diverse social media platforms, how to ensure models remain up to date when the nature online conversations is rapidly changing over time (Florio, Basile, Polignano, Basile, & Patti, 2020; Laaksonen, Haapoja, Kinnunen, Nelimarkka, & Pöyhtäri, 2020), and how to differentiate between the most egregious extreme digital speech and more moderate, but still offensive, language.

In this work we present a novel approach to address these challenges. Using data from across four different social media platforms, combined with an ordinal approach to rank extreme digital speech subtypes, and context-aware semantic embeddings, we build a model for hate speech detection which is more adaptable and comprehensive when compared to training models on data from a single social media platform in isolation. We assess the validity of our approach in three ways. Firstly, we compare performance of our cross-platform approach to models trained on data from a single social media platform to measure improvements in performance gained from leveraging cross-platform information. Secondly, we compare these performance measures to those from the existing literature for identical or similar datasets. Finally, we compare performance on a completely new and unseen dataset from a different social media platform, Reddit, and compare the performance of our proposed approach to the existing 'state-of-the-art' model trained on data from a single social media platform.

## Defining online hate speech

Hate speech is a form of dangerous online communication or 'extreme digital speech' (Ganesh & Bright, 2020), which dehumanises its victims according to their group identity, and amplifies the group identity of the perpetrator by attempting to create an antagonistic relationship between ingroup and outgroup members (Pohjonen, 2018). This 'othering' serves to reinforce existing group boundaries or create new ones. Given this potential for hate speech to drive intergroup conflict, many attempts have been made to formally define it, however as yet there is no single agreed upon definition. Various legal definitions of hate speech highlight the requirement for hate speech to cause harm (Delgado, 1982), and that this harm may include the incitement of further hostile actions beyond the speech itself (Marwick & Miller, 2014), whilst serving no redeeming purpose (Ward, 1998). These traits are not definitive however, and a wide array of different forms of speech could fit within a definition of “hate speech,” depending on the context (Sellars, 2016). A related challenge is that hate speech is not the only type of extreme digital speech, instead it sits in the middle of a spectrum of online abuse that ranges from microaggressions, condescension, and insults at the lower end, up to the promotion of self-harm, incitement of violence, and physical threats at the higher end (Jurgens, Chandrasekharan, & Hemphill, 2020). How to efficiently differentiate between elements at the lower and higher end remains an ongoing effort.

One of the few areas of consensus separating hate speech from other forms of harmful speech, is that hate speech targets groups or individuals as they relate to a group, compared to person-directed abuse which is focused solely on personal characteristics (Sellars, 2016). In this way, broad categorisations of hate speech typically present it to be bias-motivated, hostile, and malicious language targeted at a person or group because of their actual or perceived innate characteristics. Offensive language which expresses incivility to an individual, but not a group, would sit below hate and systemic intolerance towards a targeted group on this spectrum of online abuse.

In this work we follow this group-level approach, and use a definition of hate speech as “*messages which express hatred towards a targeted group with the intention to be derogatory, to humiliate, or to insult members of that group*” (Davidson, Warmsley, Macy, & Weber, 2017). Specifically, we make of use the definition proposed by de Gibert, Perez, García-Pablos, & Cuadros, (2019), and expand it slightly to incorporate implicit and indirect attacks. This definition requires three features for a message to constitute hate speech: the message must be (1) a deliberate attack, (2) directed towards, or about, a specific group of people, and (3) motivated by, or focused on, aspects of the group’s identity. Importantly, this definition is broad enough to contain any targeted group and we do not differentiate between protected and wider group level characteristics.

## Impacts of hate speech

Online hate speech is also often linked with offline hate crime. Violence towards immigrants within western countries is related to the degree of extreme digital speech and anti-refugee sentiment expressed on social media in areas where the violence takes place (Müller & Schwarz, 2017), while in the US, anti-Muslim messages disseminated by President Trump over social media correlate with the number of anti-Muslim hate crimes in states where social media usage is high (Müller & Schwarz, 2018). In addition, there is evidence for a temporal and spatial association between online race and religion motivated hate speech, and offline racially and religiously aggravated crimes (Williams, Burnap, Javed, Liu, & Ozalp, 2019). Finally, antagonistic discussions between opposing groups members has been shown to be predictive of offline violence between these same groups later in time (Gallacher, Heerdink, & Hewstone, 2020). The impact of online hate speech is also increasingly felt globally. The UN has highlighted that hate speech on Facebook played a leading role in inciting genocide of the Rohingya community in Myanmar (Stecklow, 2018), and in Sri Lanka anti-Islam hate speech has been linked with deadly mob violence (Samaratunge & Hattotuwa, 2014). Increases in the use of hate speech have also been observed after terror attacks and in particular increases in posts that advocate violence amongst Islamic extremist communities online (Olteanu, Castillo, Boy, & Varshney, 2018).

Online hate speech also causes psychological impacts even when offline violence does not occur. Following receiving online abuse people can feel afraid to leave their homes after (Awan & Zempi, 2016) along with reports of greater feelings of fear, anger, sadness, depression, as well as an increase in animosity and prejudice against the attacker’s own group (UK Safer Internet Centre, 2016; Williams & Mishcon de Reya, 2019). In addition, users who display dehumanising language and extreme digital speech also express their desire for psychological and physical separation from the out-group, which increases both ingroup identity and also affective polarisation with the out-group (Harel, Jameson, & Maoz, 2020). Understanding the nature of online hate speech is therefore important in developing any mitigation measures or policy responses.

## Current state of the art for automated hate speech detection

Early approaches to automatically detect online hate speech used lexicons and bags-of-words approaches to identify terms associated with hate speech (Warner & Hirschberg, 2012). Performance for these approaches is poor however, and they suffer from missing the context a message is posted in, and thus often miss the subtle or indirect ways in which language is used and lead to large number of false negatives, or even false positives when offensive terms are re-purposed or reclaimed for benign and even positive purposes (Gitari, Zuping, Damien, & Long, 2015; Greevy & Smeaton, 2004). Recently, performance of hate speech detection models has improved due to advances in natural language processing, general-purpose language models, machine learning, and statistical modelling (Vidgen, Tromble, et al., 2019). Key innovations include the use of deep learning and ensemble model architectures, using contextual word embeddings from pre-trained natural language models, and the inclusion of user-level variables (Badjatiya, Gupta, Gupta, & Varma, 2017; Zhang & Luo, 2018). Recent approaches using these techniques have shown promise (Fortuna & Nunes, 2018; Pitsilis, Ramampiaro, & Langseth, 2018; Rizoiu, Wang, Ferraro, & Suominen, 2019). Similarly, the inclusion of parts-of-speech tagging, especially typed dependencies and detection of linguistic 'othering' is shown to improve performance of hate speech detection (Alorainy, Burnap, Liu, & Williams, 2018; Burnap & Williams, 2015, 2016). In addition to classification at the message level, extreme users who spread hate speech can also be automatically detected using machine learning approaches (Fernandez, Asif, & Alani, 2018; Ribeiro, Calais, Santos, Almeida, & Meira, 2018).

## Limitations of these existing approaches and proposed solutions

Despite these recent improvements, limitations remain and a robust, widely applicable automated method to detect online abusive content remains to be devised (Vidgen, Tromble, et al., 2019). A key issue stems from data sparsity and lack of variability in training data (Schmidt & Wiegand, 2017), which means that classifiers typically struggle to perform well on unexpected inputs or in novel contexts. Three specific key challenges relate and follow from this; (i) how to build classifiers which work across different social media platforms, (ii) how to distinguish between hate speech and other less severe forms of extreme digital speech such as offensive or uncivil language, and (iii) how to keep classifiers up to date in a constantly changing online environment. Here we tackle these challenges with a novel approach leveraging cross-platform datasets, pre-trained natural language models, and an ordinal approach to distinguish hate speech from uncivil/offensive language.

<u>i) Cross-platform classification</u>

The first significant challenge is how to apply hate speech detection systems in a cross-platform manner. Hate speech exists on multiple social media platforms to varying degrees, but it is presented differently on each platform both in terms of structure and content. For example, the presentation of hate speech on a platform such as Twitter, where messages are short and fairly self-contained, will be very different to that on a static webpage, forum, or news website where the comments are posted in response to an article. A successful approach should be able to detect hate speech on multiple platforms without bias.

Most existing methods focus on a single platform, typically Twitter, due to its accessibility (Vidgen & Derczynski, 2020) and have relatively small training sets. These models therefore do not scale well to other platforms (Schmidt & Wiegand, 2017; Cihon & Yasseri, 2016), and risk biasing datasets towards certain types of abuse and framing (Gröndahl, Pajola, Juuti, Conti, & Asokan, 2018). Combining datasets from multiple platforms is promising (Vidgen & Derczynski, 2020), but cross-platform approaches are scarce and typically join data together in a single cohesive dataset (Corazza, Menini, Cabrio, Tonelli, & Villata, 2019; Karan & Šnajder, 2018; Mishra, Yannakoudakis, & Shutova, 2019; Salminen et al., 2020; Sigurbergsson & Derczynski, 2019), which could lead to interference between datasets as platform-specific nuances may be lost.

We address this limitation by using a 'superlearner' approach. Superlearning is a technique for prediction that involves combining many individual machine learning models into a single prediction (van der Laan, Polley, & Hubbard, 2007). In this way, individual platform-specific supervised models are first trained on datasets from four social media platforms representing a wide range of mainstream and fringe platforms commonly used by online hate groups (Twitter, Facebook, Gab and Stormfront). Then, a superlearner classifier is trained on the predictions of all four models to learn the optimal weighted average of predictions in different situations. This allows us to improve performance by leveraging cross-platform datasets without destructive interference or the need to build large models from all platform data combined. Such ensemble approaches outperform a single algorithm in isolation (Kennedy, 2017; Polley & van der Laan, 2010; van der Laan, Polley, & Hubbard, 2007). When making a prediction on new data, we combine predictions from platform-specific classifiers as features fed into the superlearner. This approach therefore retains platform-specific features and idiosyncrasies.

ii) Spectrum of extreme digital speech

An ongoing challenge in automated detection methods is how to efficiently differentiate between offensive language and hate speech. This is particularly difficult to solve as much of the language in both cases uses similar words but with different meanings (Davidson et al., 2017; Rossini, 2019). Attempts to address this distinction have used multi-class classifiers, with three distinct categorical classes for 'clean', 'offensive', and 'hate speech' (Davidson et al., 2017). This approach improves performance by reducing the conflation in the model between hate speech and offensive language but fails to consider these classes as a continuous spectrum of abuse and instead treats them as distinct unordered categories.

Here, we distinguish hate speech from less severe abuse with a three-class ordinal approach, using the ordinal nature of these classes to improve classifier performance. Our approach aims to detect hate speech by using the offensive category as a barrier to improve distinction between 'clean' and 'hate' classes.

Furthermore, distinctions between profanity and hate speech are often unclear from surface-level n-grams and heavily rely on context (Zampieri et al., 2019). As such, we use contextual word embeddings—which consider the meaning of words depending on their context—to distinguish between hate speech and offensive language. These can help detect coded language or euphemistic hate terms which traditional approaches would miss (Magu & Luo, 2018). We use embeddings from Google's BERT model, which have previously been successful in detecting hate speech (Kennedy et al., 2020; Salminen et al., 2020).

iii) Model updating

The final challenge is how to keep hate speech detection models up to date as both the nature of the language used online, and the social media platforms themselves, change over time. The specific terms used in hate speech and use of euphemistic language has been to shown change rapidly online (Florio et al., 2020; Laaksonen et al., 2020) and models will need to be updated regularly to account for this. Without regular updating machine learning model performance degrades steadily (Zliobaite,

Pechenizkiy, & Gama, 2016). Equally, in practice, when new labelled datasets for hate speech detection become available in the research community it should be possible to update existing models to utilise this new data without the requirement to retain the entire model, which would be both computationally expensive and risk destructive interference with the previous models.

We address this limitation with a two-step approach which allows for the inclusion future datasets into the model in a way which does not involve retraining the entire model. Instead, we propose to train a smaller model on the specific new dataset and then update the superlearner stage to include this new model alongside the existing ones. We discuss how this approach could help address the challenge of model degradation and updating.

# Methods

Here we propose an approach to improve hate speech detection in English language social media posts by leveraging cross-platform datasets of training data from Facebook, Gab, Stormfront and Twitter. We combine state-of-the-art natural language representations with syntactic features and an ordinal approach to classify hate and offensive speech to improve classifier performance.

## Hate speech and offensive language definitions

As outlined in the introduction, we define hate speech as messages which contain (1) a deliberate attack, (2) directed towards, or about, a specific group of people, and (3) motivated by, or focused on, aspects of the group's identity. Importantly, this definition is broad enough to contain any targeted group, unlike some legal definitions and the definitions used by social media platforms which focus only on explicitly protected groups and specific characteristics (Facebook, 2020). This places our definition more in line with 2020 UK Law Commission recommendation, which gives an 'other' category for group membership in addition to explicitly named groups with protected characteristics (Law Commission, 2020).

Offensive posts are broadly defined as messages likely or with the intention to cause offence (Davidson et al., 2017), this includes uncivil, rude, inappropriate, or overly disrespectful content. It often contains obscenity and profanity towards a recipient in the 2nd or 3rd person but without invoking aspects of the groups' identity as motivation for the attack. It also includes derogatory language (e.g. sexist or homophobic language) to cause offence, but which does not attack the group because of one of its characteristics. This distinction is common in the literature (Vidgen, Tromble, et al., 2019; Waseem, Davidson, Warmsley, & Weber, 2017; Zampieri et al., 2019). As an example, attacking an individual by using a slur but with no other referral to aspects of the individual's group identity would be classified as offensive (e.g. 'you're such a b***h'). Comparatively if a phrase derogated a group as a whole (e.g. 'all journalists are untrustworthy tw*ts') or called for hostile actions against this group (e.g. 'all journalists should be imprisoned') then this would be classified as hate speech.

Importantly, we do not distinguish 'sarcastic' from 'genuine' hate or abuse and classify this content as if it were serious. The high degree of sarcasm, irony, and in-jokes present in these platforms (Zannettou, Caulfield, et al., 2018) has been shown to provide a source of classification error as this distinction is increasingly blurred (Vidgen et al, 2019), especially as this content usually relies on the tacit acceptance of the prejudice and negative tropes even when displayed in a humorous way (Ma, 2014).

## Training datasets

We use data from four key social media platforms popular among the far-right: Facebook, Twitter, Gab and Stormfront. Facebook and Twitter are popular with users from across the political spectrum but contain conversations from extremist groups who spread large volumes of hate speech (Burke, 2017; Vidgen, Yasseri, & Margetts, 2019). Gab and white supremacist forum Stormfront are more fringe but popular with extremist organisations and contain considerable hate speech (Kleinberg, Vegt, & Gill, 2020; Mathew, Dutt, Goyal, & Mukherjee, 2019). We selected these platforms because they vary substantially in their structure, the length of messages that can be posted, and specific groups which use them, therefore allowing us to gather a breadth of content and build a nuanced classifier. In total we collated a dataset of ~40,000 messages to build our classifier, split approximately equally across the platforms (Figure 1, Supplementary Information (SI) Table 1).

For Facebook we take a random sample of 10,000 messages from a dataset previously analysed in the context of intergroup conflict (Gallacher et al., 2020). This contains messages posted to 20 event pages from 2015 – 2017 created by two far-right groups in the UK: Britain First and the English Defence League. Both groups have subsequently been banned from the platform for expressing hate speech and spreading extremism.

For Gab we take a random sample of 10,000 messages from across the entire platform posted between its formation on 10th August 2016 and the 29th October 2018, the date at which it was taken temporarily offline. Our data consists of an amalgamation of data shared by Zannettou, Bradlyn, & Cristofaro, (2018) (August 2016 – January 2018) and data from the online repository Pushshift (January 2018 – October 29th 2018).

For Twitter we use a dataset published by Davidson et al. (2017) containing a sample of 24,802 tweets from users who expressed words from a lexicon of hate terms and already coded into "hate speech", "offensive" and "clean" classes (details below). We subsample 10,000 messages from this dataset by randomly down sampling the majority classes.

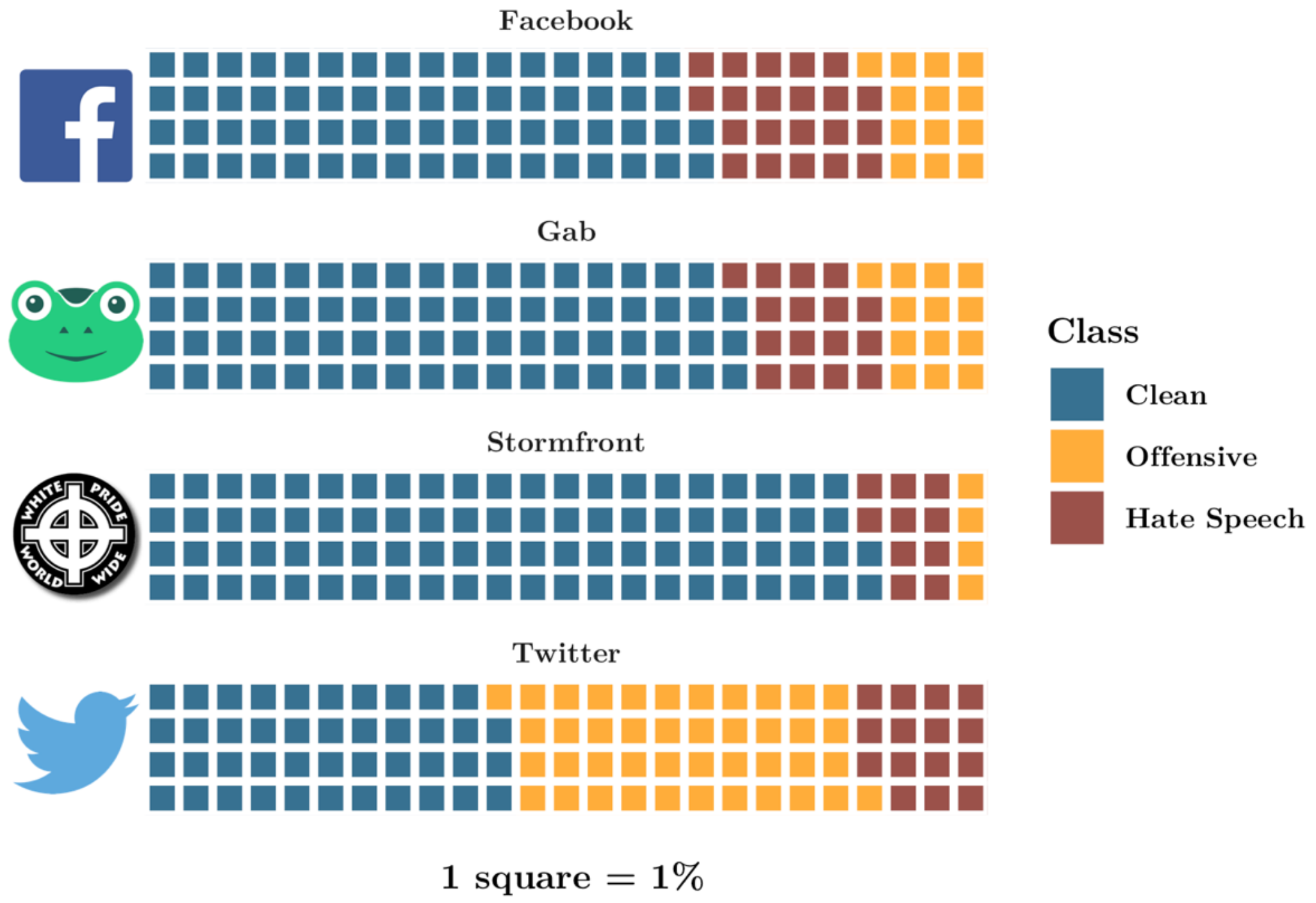

*Figure 1 - Proportion of Hate Speech, Offensive Language and Clean Messages in the training data from across the four social media platforms*

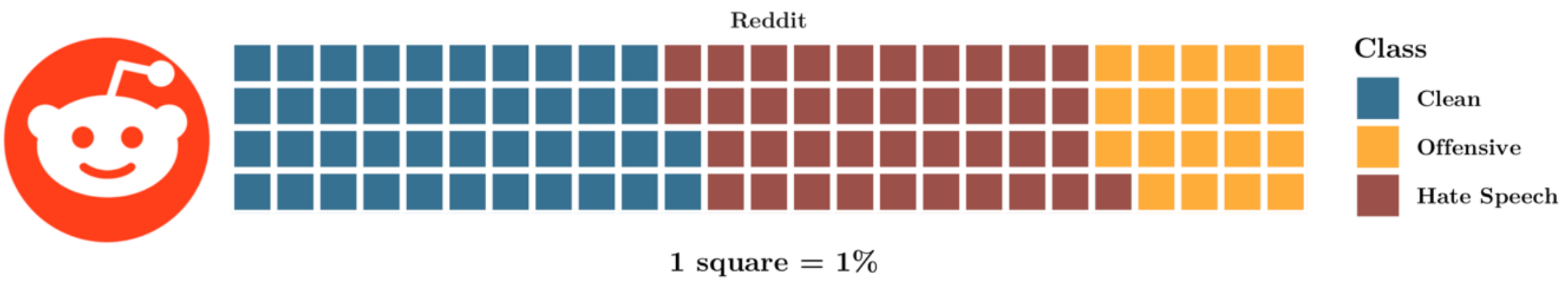

*Figure 2 - Proportion of Hate Speech, Offensive Language and Clean Messages for the Reddit validation set*

Finally, we use a dataset containing 10,568 messages from the white supremacist forum Stormfront made publicly available by de de Gibert et al., (2018). This dataset contains a random sample of English language posts from any of the 22-sub forums of the platform between 2002 and 2017 and is already classified into "hate" and "non-hate" messages.

The training data from the four platforms contained different proportions of messages from each class, as shown in Figure 1. On all platforms, clean messages formed the dominant class, however while Facebook and Gab datasets contained a fairly even balance of hate speech and offensive messages as minority classes, this was not the case on Twitter and Stormfront, with Stromfront having few offensive messages compared to hate speech, and Twitter having a lot more. This reflects both the sampling strategies used in the collection of this data, with Twitter data sampled using keyword searches which boosts the presence of offensive language for example, and also the underlying platform dynamics, with Stormfront containing less offensive messages relative to hate speech. This does not mean however that Stormfront should be considered a 'cleaner' platform, as this doesn't account for the severity or impact of the hate speech when it occurs on the platform.

## Novel platform validation dataset

To test the performance of our modelling approach on a novel social media platform not included in the original training, we use a random sample of 1,000 annotated posts from a different social media platform, Reddit, shared by Qian, Bethke, Liu, Belding, & Wang, (2020). These messages were pre-labelled for presence or absence of hate speech, but we re-annotate them to fit our three categories.

We deliberately chose 1,000 posts as they represents a 'new' dataset becoming available, but of insufficient size to be used efficiently on its own, as it is well below the average for training a completely new hate speech detection model (Vidgen & Derczynski, 2020). However, if combining a model built from this 'small' dataset with the other existing platform-specific models then this would support the validity our approach. The balance of data across the three classes for the Reddit data is shown in Figure 2.

## Data labelling / annotation approach

With the exception of the Twitter dataset, already labelled into 'hate speech', 'offensive' and 'clean' categories by Davidson et al. 2017, all datasets required full (Facebook and Gab) or partial (Stormfront, Reddit) manual labelling. To ensure consistency across datasets we used the same labelling strategy as was used for the Twitter dataset, which is based on the definition of hate speech described above. We first labelled the Facebook and Gab datasets into 'hate' and 'non-hate' categories. Then, we manually split the 'non-hate' speech category into 'offensive' and 'clean'. For Stormfront and Reddit, to allow comparability in performance for hate speech detection with the existing literature, we did not re-classify any messages from within the 'hate speech' class of this dataset but created an 'offensive' class by reclassifying offensive messages previously labelled as 'non-hate'.

All data labelling was conducted by a single native English-speaking coder, familiar with the far-right online ecosystem and language used. To validate the accuracy of this labelling, a sample of 1,000 messages from Facebook and Gab platforms was selected and labelled independently by a second coder who is also familiar with the online environment but has a different nationality, gender and background, who had been trained in data labelling strategy and hate speech definition prior to annotation. Intercoder reliability scores gave a percentage agreement of 89.7%, an ordinal Krippendorff's Alpha of 0.863, and a Cohen's Kappa score of 0.797. The former is at the level of 'good' agreement (George & Mallery, 2003) while the latter is well above the 0.61 threshold for reliable coding and substantial agreement (Glen, 2014), and so we retain our labelling as accurate.

## Data pre-processing

After manual labelling, data were pre-processed prior to building the machine learning models. We removed any messages that had fewer than 2 words or fewer than 5 characters in total, as these were too short to extract meaningful features, or which could not be classified by the human coders into a category. These messages were uncommon and included examples such as '*it is*' or '*they are*' which cannot be meaningfully classified, and presumably rely heavily on context not included in the message itself. In this step we also removed messages in languages other than English and any messages which contained no substantial text (e.g. only URLs or images). We aimed to build a classifier looking at messages in isolation, so we filtered out the 1,001 comments from the Stormfront dataset which had been coded by de Gibert et al. as requiring additional context for human annotators to code – i.e. they could only be labelled by looking back within the comment thread to identify the context of the conversation. Messages which could be classified in isolation, even if embedded in a thread, were still included. This therefore matched with the approach taken on Facebook and Gab labelling, where messages which could not be coded in isolation were removed. The final samples sizes for each platform is given in the Supplementary Information (SI) Table 1.

Following this, all messages were put into lowercase and any non-standard (UTF-8 / ASCII) characters removed. We used regular expressions to extract the title of linked pages or news articles from messages containing URLs in addition to other text (see SI 1.2). The remainder of the URL was discarded. All punctuation, including hashtags, was counted and then removed, although the keyword following a hashtag was retained and treated like any other word within the message.

## Features used in machine learning classification

We extracted commonly used and novel features from each message to train the model to classify them into 'clean', 'offensive' or 'hate speech' categories. Broadly, these features can be broken down into semantic features that encode the substance of the message, and syntactic features that encode how the message is presented. In total we extracted 7 types of semantic feature and 6 syntactic features. These are listed in SI Table 2.

<u>Semantic features</u>

For the majority of the semantic features we use word embeddings—a learnt representation of text based on words' meaning but also the context of the entire message—to extract meaning from the text. Specifically, we make use of the Bidirectional Encoder Representations from Transformers (BERT) model (Devlin, Chang, Lee, & Toutanova, 2018) which has been trained on large quantities of unannotated online text, and utilises bidirectional search (forward and backward from every given word) to better understand the use of a word in the context of the whole message. We extracted BERT document vectors (distributed representations of the entire text across multidimensional space) for each message using the 'rBERT' package (Harmon & Bratt, 2020) and took the final layer output vector for the "CLS" token. This creates a representation of each message in 768 features. Extracting BERT word vectors and feeding these forwards in downstream models performs at near parity with fully retraining the model (Alammar, 2018).

In addition, we encode plural nouns, 'othering' and explicit hate terms; theory-driven semantic features likely to be present with hate speech expressions. Plural nouns may be especially informative for hate speech classification as they may indicate when a message is directed towards an outgroup as a whole – a necessary element of hate speech in our definition. This alone may not indicate hate speech but may be a good indicator when combined with other features such as the content of the message or the presence of offensive language. We used a part of speech (POS) tagging system to extract all plural nouns and identify their relevance for each label category by calculating the weighted-log-odds (details in SI 1.4). Derogatory and pejorative plural nouns, as well as standard terms for groups often receiving abuse, received high scores within the hate speech category; in contrast top terms in the 'clean' category are not offensive (SI Figure 1).

We measure the presence of 'othering' by identifying the use of two-sided pronouns that contain a distinction between the in-group and out-group in a single message (e.g. your/our, them/us, they/we), using dictionaries of popular personal pronouns for both the in-group and out-group. This can improve hate speech detection, especially when the group being targeted is not explicitly named (Alorainy et al., 2018).

Additionally, we measured the presence and severity of explicit hate words/symbols using a crowdsourced dictionary of 883 'hate terms' rated by severity on a 1-100 scale (Quinn, Tuckwood, & Boyd, 2019). For each message we calculated the number and cumulative hate severity score of these hate terms. Furthermore, as extremist groups are increasingly using hate symbols to express coded hate we included a measure of the number of hate symbols in a message, using a dictionary of hate symbols (Anti-Defamation League, 2019), (e.g. '(((echo)))' is used by alt-right online communities to target Jewish individuals).

Additional features included overt obscenity, counted as the number of swear words per message using an online dictionary used to block profanity in online message boards (Fontgate Media, 2014), and the sentiment / polarity of the message on a -1 to +1 scale using the polarity function from the 'qdap' package (Rinker, 2020), as both hate speech and offensive language are likely to have a more negative valence.

<u>Syntactic features</u>

Syntactic features include counts of the number of words, sentences, punctuation, and pronouns, plus the presence of negation. We also calculated the complexity of the messages measured as the lexical density (the ratio of content words to total words (Halliday, 1989) and their readability, measured with Flesch-Kincade reading ease scores (Kincaid, Fishburne, Rogers, & Chissom, 1975). These have both been shown to be predictive features of online hate speech in certain contexts (Zhang, Robinson, & Tepper, 2018), while more complex language is inversely related to extreme out-group derogation (Park & DeShon, 2018; Zmigrod, Rentfrow, & Robbins, 2019).

## Machine learning approach - modelling

Our modelling takes a two-stage approach. We first built four separate platform-specific classifiers, one for each social media platform. Once these four models were trained and the hyper parameters tuned, we then combined them using a superlearner approach to improve performance while retaining platform-specific characteristics (Figure 3).

Step 1- Platform-specific classifiers

For each platform we built two binary classifiers, one trained on the labels 'hate speech' vs 'not hate speech' (classifier 1) and another trained on 'clean' vs 'not clean' (classifier 2). We then combined the outcomes of these two models using the formula below to create a single ordinal classification ranked by increasing severity from 'clean' to 'offensive' to 'hate speech. For ordered classes this approach outperforms the naïve approach which treats the classes as an unordered set, and can be applied to any machine learning algorithm (Frank & Hall, 2001). Where *p(clean)* and *p(not clean)* are obtained from classifier 1 and *p(hate speech)* and *p(not hate speech)* are obtained from classifier 2.

Clean (lowest severity) : *p (clean) = 1 - p (not clean)*
Offensive (intermediate severity): *p (offensive) = p (not clean) – p (hate speech)*
Hate speech (highest severity*): p (hate speech) = 1 – p (not hate speech)*

We split the data into a training and a test set (80% and 20%) for each platform, stratified by categories to ensure balance to the 'real-world' (full) dataset. This test set was then held out for final model verification. The modelling validation was done using 10-fold cross validation (in addition to the 20% held out sample for final model verification), again with samples stratified on message category within each fold. In order to avoid information leakage between training and validation steps we included a number of measures within the cross-validation process. In each fold the data was down-sampled to a ratio of 2:1 from the majority class to the minority class using the original classes (Hate/Offensive/Clean) to truly reflect the underlying data. All predictors were then centred, scaled, and all the near zero variance predictors removed. The feature space of BERT word vectors was reduced using a supervised partial least squares (PLS) approach to maximise the separation of principle features between classes (Kuhn & Johnson, 2013), which has been shown to improve performance of word vector models (Gupta, Giesselbach, Rüping, & Bauckhage, 2019; Raunak, Gupta,

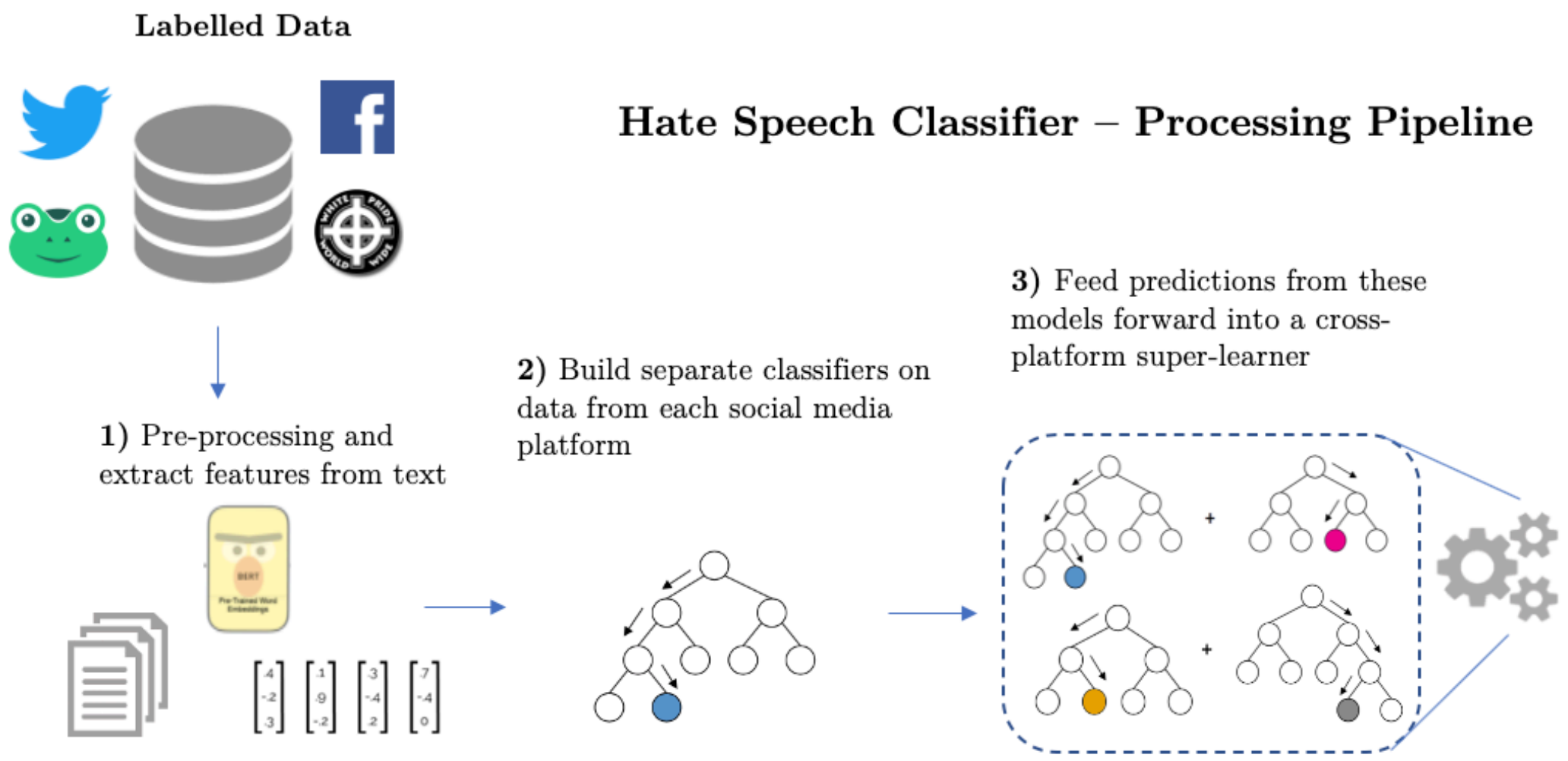

*Figure 3 - Modelling pipeline for hate speech detection and type of hate classification*

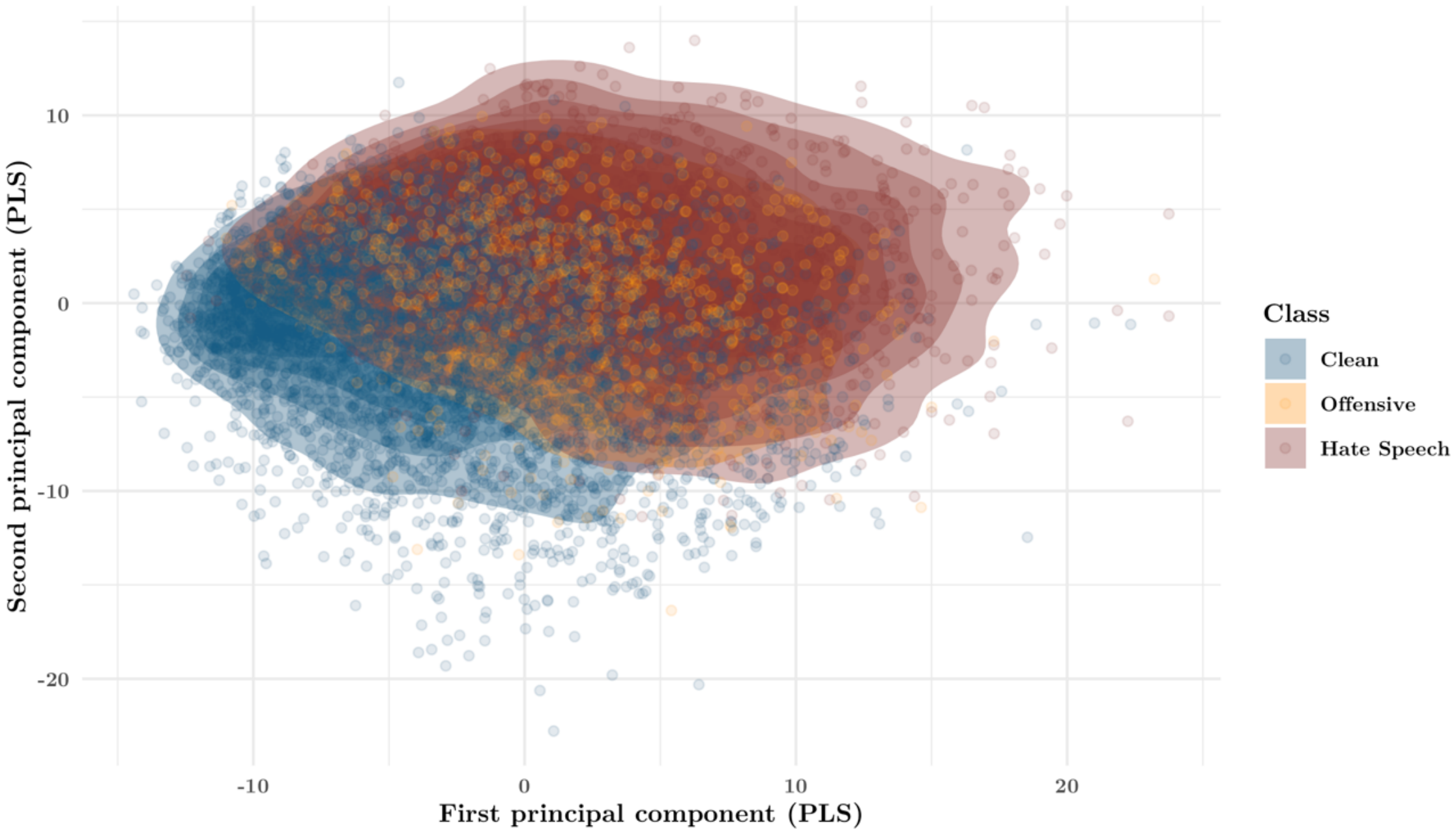

*Figure 4 – Partial Least Squares (PLS) dimensionality reduction showing a separation in contextual word embeddings across the three message categories*

& Metze, 2017). Using this PLS approach, we find that the vector space for the primary principle components populated by messages from across these three classes is different, but overlapping, suggesting that the semantic embeddings of the classes should offer some level of discrimination, especially in combination more components and other features (Figure 4).

We tested five types of models - C50 decision trees, general linear models, support vector machines, neural networks, and gradient boosted decision trees (xgboost). We found that the latter, xgboost (Chen & Guestrin, 2016), was the best performing model on all datasets and so retained this type of model for the remainder of the platform-specific analysis. For details on hyperparameter tuning, cross validation and equivalence zones see SI 1.5. Feature distinction for both semantic and syntactic features is shown in Figure SI 2.

Step 2 - Cross-platform superlearner

In training the superlearner we tested the same five machine learning algorithms again. This time we selected neural networks as they performed best in cross-fold validation. The superlearner takes platform-specific predictions from the models in step 1 as features and builds a new model by learning the optimum combination of these predictions to maximise performance. In addition to the predictions across the three classes for the four platform-specific models, each message's original platform was also included as a feature, giving a total of 13 features for the model to learn from.

To avoid the need to retain a further dataset for training the superlearner, whilst also avoiding information leakage between training and validation sets, we used the cross-validation predictions (rather than full predictions) from the platform-specific model corresponding to the original platform of the training data. We used predictions from the three other full platform-specific models. For example, for a Gab message in the training dataset of the superlearner, the predictions which serve as features were those from the full Facebook, Twitter and Stormfront models and the cross-validation Gab model. This means a model was never trained and tested on the same data points

### Classifier validation on novel data

To compare our superlearner approach to platform-specific and existing 'state of the art' approaches, we made predictions on a new random sample of 1,000 annotated posts from a different platform, Reddit. We tested the prediction performance of four different models on this Reddit dataset.

Firstly, for comparison to the existing literature, we got predictions using 'HateSonar' (Nakayama, 2018), a Python implementation of a three-class model trained solely on Twitter data (Davidson et al., 2017). This provides a baseline performance for applying hate speech detection models to unseen data from a novel social media platform.

We then trained our own platform-specific 'Reddit' model, created using the same approach as for our four platform-specific models described above, giving a baseline for models trained on a small dataset of platform-specific data.

Following this we then got predictions for two superlearner models. Firstly, we used our original superlearner model described above (superlearner 1.0), i.e. trained on data from four social media platforms but not on Reddit data. We then created an updated superlearner (superlearner 2.0) trained on the outputs of all five platform-specific models, including the new Reddit model described above. This allowed us to verify the performance improvements provided by this cross-platform approach, as well as the opportunity it provides for model updating as new data becomes available. Both the Reddit-only model and the superlearner 2.0 models were created using a 60% training set with a 40% test set held out for performance measurements. We selected a larger percentage test set for this Reddit data, compared to prior datasets, in order to ensure valid results given the smaller dataset size.

### Data analysis

All machine learning and analysis were done in R (version 3.6.1) using the tidyverse (Wickham et al., 2019) and tidymodels (Kuhn, Wickham, & RStudio, 2020) collections of packages. Where particular additional packages have been used, they have been referenced in the text.

### Ethics

All research was conducted in accordance with the University of Oxford Ethics Committee (Ethics Reference: SSH_OII_CIA_19_062). All data collection was conducted using open-source methods and publicly available data, and hence, informed consent was not explicitly obtained.

# Results

To assess the validity of our approach, we first compare the performance of the cross-platform superlearner to that of platform-specific models. Secondly, we compare these performance measures to those from the existing literature for identical or similar datasets. Finally, we compare the performance of our superlearner model on a completely new and unseen dataset from a different social media platform to an existing alternative state-of-the-art hate speech detection model.

## Superlearner vs platform-specific models

The superlearner approach correctly labelled 91.4% of the messages across all four datasets, with an F1 score of 0.69 for hate speech detection (precision 0.73, recall 0.66), an F1 score of 0.96 for detection of clean messages (precision 0.95, recall 0.97), and an F1 score of 0.86 for offensive messages (precision 0.86, recall 0.86). The performance figures and confidence intervals from across all classes, platforms, and models are given in SI Table 3 (See SI 2.1 for confusion matrix and ROC curves).

Table 2 gives the classifier performance specifically for hate speech detection for models trained on individual platforms and the superlearner performance across all four platforms, along with the relative change in performance that the superlearner approach achieved. The superlearner implementation improved overall classifier accuracy by 1.6 percentage points compared to the individual platform-specific models, and this improvement was largest on Stormfront (3.4 percentage points) and Gab (1.2 percentage points). Specifically, this approach increased the precision of hate speech classification by 0.05 at the expense of recall which decreased by 0.03 (see SI 2.2 for exploration of minor/major classification errors). In other words, this suggests that the superlearner model performed better at detecting hate speech than platform-specific models.

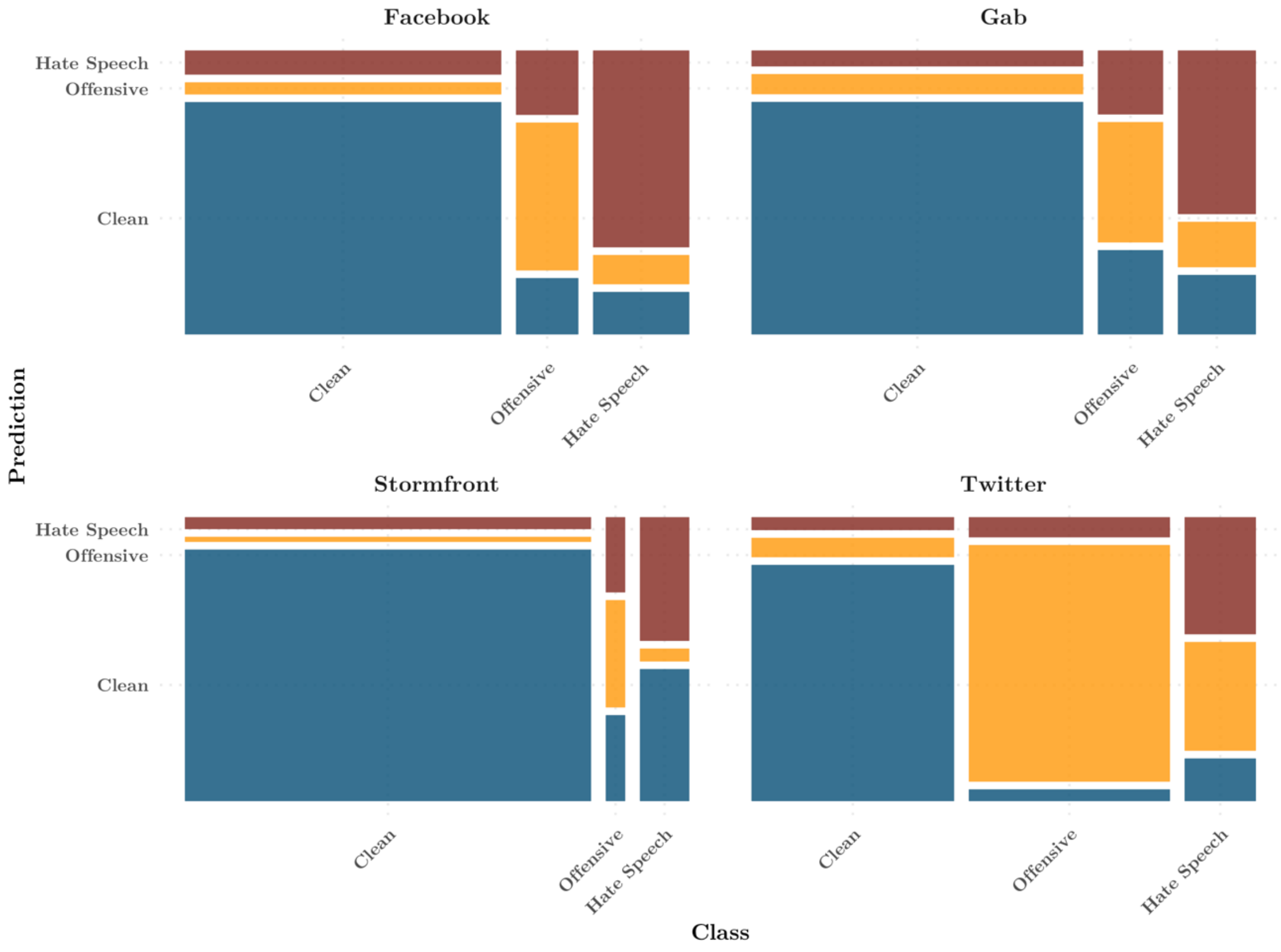

*Figure 5 – Mosaic plots for performance on different classes for the superlearner across the four platforms. Colours related to data considered (x-axis) or classified (y-axis) as 'clean (blue), 'offensive' (yellow) or 'hate speech' (red).*

*Table 2 – Classifier performance for models trained on individual platforms and the superlearner*

| | Platform-Specific Model Performance | | | Superlearner Performance | | | | Superlearner Change | | |
|---|---|---|---|---|---|---|---|---|---|---|
| ***Platform*** | **Accuracy** | **HS Precision** | **HS Recall** | **Accuracy** | **HS Precision** | **HS Recall** | | **Accuracy** | **HS Precision** | **HS Recall** |
| ***Overall*** | 89.8 | 0.68 | 0.69 | 91.4 | 0.73 | 0.66 | | +1.6 | +0.05 | -0.03 |
| ***Facebook*** | 89.3 | 0.75 | 0.83 | 90.3 | 0.75 | 0.87 | | + 1.0 | 0 | + 0.04 |
| ***Gab*** | 89.8 | 0.73 | 0.75 | 90.9 | 0.75 | 0.83 | | + 1.2 | + 0.02 | + 0.08 |
| ***Twitter*** | 89.6 | 0.76 | 0.40 | 89.9 | 0.71 | 0.36 | | + 0.3 | - 0.05 | - 0.04 |
| ***Stormfront*** | 90.5 | 0.50 | 0.77 | 93.9 | 0.67 | 0.42 | | + 3.4 | + 0.18 | - 0.35 |

When inspecting the performance of platform specific models (Figure 5, SI Table 3) it is apparent that all four models performed well in correctly identifying the largest class of clean messages (F1 range 0.94 to 0.95, shown in blue). Performance was more mixed however when presented with offensive messages (yellow), with a higher proportion of these being incorrectly classified as either clean or hate speech when compared to predictions for clean messages (F1 scores for offensive classifications: Facebook 0.71, Gab 0.61, Stormfront 0.51). The exception to this is the Twitter model, which performed well when presented with offensive messages (Twitter offensive F1 score: 0.91), reflecting the higher proportion of messages from this class in the training set. With regards to hate speech (red) there were again difference in performance across the platforms, with Facebook and Gab models correctly recalling a higher proportion of hate messages (0.83 and 0.75 respectively), whereas the Twitter model failed to detect a much larger proportion of these messages (recall 0.40), and the Stormfront model achieved lower precision in hate speech classifications (precision 0.50). Interestingly, classification errors were not evenly distributed across incorrect classes, and instead followed the ordinal pattern, with hate speech messages more likely to be incorrectly classified as offensive than clean. The reverse is true for clean messages, which were more likely to be incorrectly classified as offensive than hate speech. This effect is shown clearly in the mosaic plots for Facebook and Gab in Figure 5.

For platform-specific models, the results also showed that when making predictions for data from a different social media platform to the training data (for example getting predictions for the Facebook data from the Twitter model), performance was reduced (SI Table 4). While performance was above chance for all combinations of models and platforms, it was significantly lower than the performance for congruent platform/data combinations, and also lower than the superlearner performance. The distributions of hate speech predictions for one platform against the others in shown in Figure SI 5, indicating how the overlap between platform-specific hate speech varies.

## Comparison to existing approaches

Our hate speech detection model performed better in comparison to the published literature using the same or similar datasets.

For Twitter data, Davidson et al (2017), used a tf-idf bag of words approach with some additional features (lexical density, sentiment, and syntactic features) produced an overall model accuracy of 90% and precision and recall values for hate speech at 0.44 and 0.61 respectively[1]. Our approach reported similar overall accuracy performance for but improved the precision of hate speech detection substantially. Our superlearner achieved 0.71 precision for hate speech messages on Twitter (Table 2) vs the 0.44 precision for this same data reported by Davidson et al, therefore reducing the presence of false positives.

For Gab data, a similar approach by Kennedy et al, (2020) using a model trained with BERT word embeddings reported an overall classifier accuracy of 87%, with precision and recall values for hate speech detection at 0.59 and 0.57 respectively. Our superlearner approach therefore improved both precision and recall substantially, reducing both false positives and false negatives for hate speech detection by approximately 20 percentage points.

For the Stormfront data, a recurrent neural network with short-term memory (LSTM) produces an overall accuracy of 78% under similar conditions (de Gibert et al., 2018), which is 16 percentage points lower than the accuracy our superlearner achieved. Precision and recall values are not available for comparison. There is no comparable model available for Facebook, however Facebook themselves reported in May 2020 that they detect 89% of hate speech messages posted to the platform using automated techniques (Rosen, 2020), which is broadly in line with our observed performance.

## Performance on data from a new platform

In order to investigate how well this superlearner approach (superlearner 1.0) performed when tested on data from a novel social media platform not included in the original training sets, we measured its performance on a set of messages from Reddit, and compared this performance to an existing model (HateSonar) trained using data from just a single social media platform. For completeness, we also trained a new model using this Reddit data and included performance from this model in our comparisons, as well as from a superlearner updated with this new Reddit specific model (superlearner 2.0).

When testing performance on data from a novel social media platform not included in the original training set, we find that all three of our models (Reddit-only, original superlearner 1.0 and updated superlearner 2.0) performed substantially better than the HateSonar implementation of Davidson et al's approach (Table 3). This is particularly pronounced in the recall of hate speech messages, where the HateSonar model detected only 18% of the hate speech messages in the Reddit dataset while both of the superlearner approaches correctly detected 87% of these same messages – a substantial improvement.

The superlearner 1.0 model and newly trained platform-specific model achieved similar performance, with the platform-specific model scoring a higher overall accuracy but the superlearner 1.0 performing better on hate speech precision and recall, overall increasing the F1 score for hate speech detection by 0.02. The best performance however was obtained by combining the newly created Reddit-model with the pre-trained models from the other four platforms via the superlearner approach. This updated superlearner approach (superlearner 2.0) achieved an overall accuracy of 89.0% across the three classes and an F1 score of 0.92 for hate speech predictions (precision 0.96, recall 0.87). The performance figures for all classes and models on this Reddit dataset are given in SI Table 4.

---

[1]*Note – Davidson et al appear to train the final model on the entire dataset and then get predictions for these same tweets i.e. no held out test set was used. This is likely to overestimate performance, particularly around recall and this should be taken into consideration when comparing performance.*

*Table 3 – Classifier performance on new and unseen Reddit data for the HateSonar model, platform-specific classifier, and the superlearner models*

**Model Performance on Reddit Data**

| ***Model*** | **Accuracy %** | **HS Precision** | **HS Recall** | **HS F1** |
|---|---|---|---|---|
| *HateSonar* | 61.6 | 0.81 | 0.18 | 0.30 |
| *Reddit Platform-Specific* | 86.3 | 0.87 | 0.84 | 0.85 |
| *Superlearner 1.0* | 83.6 | 0.88 | 0.87 | 0.87 |
| *Superlearner 2.0* | 89.0 | 0.96 | 0.87 | 0.91 |

# Discussion

Our cross-platform approach for hate speech detection outperformed existing models, both on data similar to that which these models are trained on, and for unseen data from a novel social media platform not included in the original training set. This highlights the opportunities of cross-platform approaches to improve automatic detection of online hate speech. In addition, we found that performance on novel platform data was easily improved by adding to the superlearner a new model trained on a small dataset from this novel platform, demonstrating the flexibility and ease of updating provided by this approach.

## Benefits of the Superlearner

<u>i) Leveraging Cross Platform Information</u>

A key benefit from our approach is to combine predictions from multiple platform-specific datasets. This is found to improve performance on hate speech detection over using a data from a single social media platform in isolation. This improved performance is likely because new types of hate speech or specific formats of hate expression in the test data from a platform-specific model may be present in the training data from other platforms. Our cross-platform approach helps solve the challenge of data sparsity and lack of variability in hate speech detection training data (Schmidt & Wiegand, 2017) by leveraging datasets from across multiple social media platforms. Interestingly we find that simply applying models in an incongruent cross-platform approach without using this superlearner does not lead to good performance – platforms trained on just one social media platform do not extrapolate their predictions well to another platform without adjustment. This suggests that the nature of hate does vary significantly across platforms (which is the variability we wish to capture) but creating any cross-platform model that does take into account the platform from which a message originated will struggle to perform at a high level.

Interestingly, we find that overall this improvement in performance is driven by increased precision of predictions; in other words the superlearner is less likely to make a false positive hate speech judgement than the individual platform models, while the proportion of false negatives is increased slightly. The applicability of this approach may therefore be more useful in cases where this trade-off (fewer false positives) is preferable. This effect is platform-specific however. For Gab data for example, we observe that the superlearner improves both hate speech precision and recall; reducing both false positive and negatives.

Another benefit of our approach is that it will reduce the impact of biases or errors in training data, because the platform-specific model trained on this data will then be compared to other predictions from other datasets and models. Estimates of the Twitter data from Davidson et al (2017) suggest that up to 10% of the data is mislabelled, and this will reduce performance, especially on new out-of-sample datasets. In addition, the data has been shown to be skewed towards racism and sexism compared to other forms of hate speech (Vidgen, Tromble, et al., 2019). Similarly, the Stormfront dataset from de Gibert et al (2018) focuses primarily on white supremacy rather than other forms of hate. Using multiple disparate datasets to build models and combining predictions will reduce the impact of errors and biases in any one dataset.

ii) Future updating and expansion

The accuracy of hate speech detection models can degrade quickly as the nature of online language changes (Florio et al., 2020; Laaksonen et al., 2020). Keeping models up to date with new training data is therefore important to preserve accuracy. However, re-training the entire model is computationally expensive, and risks destructive interference if new data contains errors or differences in labelling from prior training data. Our superlearner approach is expandable without the need to retrain each single-platform model as only the new data needs to be used to train a new model. This reduces computational costs, and these predictions can easily be combined with existing predictions in a new superlearner, as we demonstrate with a new Reddit dataset. While the cross-platform superlearner (1.0) performed better than existing approaches on this dataset, training and combining a new small platform-specific model with existing pre-trained platform models into a new superlearner (2.0) noticeably improved performance. Our model can therefore be easily updated when new datasets becomes available by 'slotting' them into the existing superlearner.

iii) Ordinal distinction between offensive and hate speech

Another benefit of our approach is to include an ordinal approach to hate speech detection. Our model made fewer major category errors (misidentifying clean speech with hate speech or visa-versa) than minor category errors (SI section 2.2), which provides evidence for ordinal information being contained in features across the three classes. Training platform-specific models using this ordinal approach, with the addition of the theoretically informed features that we have included, improved performance within the Gab dataset compared to a comparable classifier containing just a binary 'hate' and 'not hate' distinction (benchmarking with Kennedy et al 2020). Future research should investigate whether a more nuanced distinction between types of extreme digital speech and the creation of more than three classes can further improve this ordinal approach.

**Limitations and future directions**

Our superlearner approach shows potential for future automated hate speech detection, but limitations remain. First, variation in hate speech definition across studies and labelled datasets (Vidgen & Derczynski, 2020) makes comparing models or combining datasets challenging; in this study we had to manually re-label data for consistency. Combining predictions into a superlearner may smooth some variation but problems may still occur. Consistency in definitions and class labels will help the field move forwards and widen the applicability of this superlearner approach.

Bias in data labelling is another limitation which the field of hate speech detection should aim to identify and reduce. Prior models of hate speech detection found, for instance, that messages in African-American English were more susceptible to being misclassified as hate speech than those in 'standard American English' (Davidson, Bhattacharya, & Weber, 2019). These racial biases need to be addressed in any machine learning system prior to deployment to avoid discrimination, e.g. with balanced training sets from across a wide range of domains and users. Biases also occur less often with data labelled by expert annotators (Sap, Card, Gabriel, Choi, & Smith, 2019; Waseem, 2016), therefore datasets labelled by expert annotators should be used whenever possible and weighed higher than crowdsourced datasets in a superlearner approach. Here we used two annotators of different genders, nationalities, and backgrounds to reduce bias in data labelling. Bias can also be introduced at the stage of extracting word vectors however (Bolukbasi, Chang, Zou, Saligrama, & Kalai, 2016). This is especially pernicious as interpretation of word vectors cannot be easily performed, and therefore the bias gets propagated forwards silently. The impact of bias in these pre-trained word vector models should be investigated further. Future work should also incorporate multiple languages as hate groups increasingly operate globally (Davey & Ebner, 2017; Sigurbergsson & Derczynski, 2019).

Automatic detection systems of extreme digital speech are also susceptible to adversarial attacks and perturbations of language (Gröndahl et al., 2018; Hosseini, Kannan, Zhang, & Poovendran, 2017), including spelling and grammar changes. This can lead to a lower judgement of hate by the classifier, but not decrease the impact on a human recipient of the message. Future research should take inspiration from the field of image recognition and include perturbations of data within the training set to help mitigate against this type of attack (Nandy, Hsu, & Lee, 2020).

Finally, while we used pre-trained general-purpose natural language models to get context-dependent word embeddings, recent improvements in training these language models on online-specific and topic-specific conversation data (Müller, Salathé, & Kummervold, 2020) may improve the context awareness and performance of the subsequent models. Pre-training BERT models on social media platforms containing hate speech is likely to improve classifier performance and is a promising avenue for future research.

# Conclusion

Overall, our approach combining datasets from across multiple social media platforms shows promise in building better automatic tools to detect online hate speech and can help address the challenges of data scarcity and low variability, and improve applicability for novel social media platforms not included in the original training data. In the context of growing threats from extremism, intergroup conflict, and online hate, developing robust and adaptable methods for automatically detecting these threats is an important step in improving online safety.

## Acknowledgements

We thank Davidson et al (2017), de Gibert et al (2018), Quian et al (2019), Zannettou et al (2018), and Pushift.io for making their data available and also the research assistants who helped annotate the datasets used in this analysis.

# Supplementary Information (SI) for:
## Leveraging cross-platform data to improve automated hate speech detection

**1 - Modelling approach**

**2 – Machine Learning Model Performance, Confusion Matrix and ROC Curves**

**3 – Incongruent cross-platform performance**

# 1 – Modelling approach

### 1.1 – Class balance and dataset size for the four platforms used in training

SI Table 1 gives the final size of the four training sets (Facebook, Gab, Twitter and Stormfront) used to build the platform-specific hate speech classifiers after pre-processing the data, along with the proportions across the three message classes.

***SI Table 1- Size of datasets used for classification and balance of classes once the data were cleaned and pre-processed***

| | | Class Balance | | |
|---|---|---|---|---|
| | **Number of Messages** | **Prop. Clean** | **Prop. Offensive** | **Prop. Hate Speech** |
| ***Facebook*** | 9,999 | 0.70 | 0.12 | 0.18 |
| ***Gab*** | 8,274 | 0.71 | 0.13 | 0.16 |
| ***Twitter*** | 9,730 | 0.43 | 0.43 | 0.14 |
| ***Stormfront*** | 9,715 | 0.87 | 0.04 | 0.10 |

## 1.2 – Extracting webpage title from URL

In the data pre-processing steps, where possible we extracted the web page title for URLs which were included in messages. This was done using regular expressions to extract text which occurred within a character string after either ‘http://’ or ‘https://’ and after the first occurring full stop and before the final forward slash. We then removed any punctuation that occurred between words in this URL. We were not able to obtain meaningful text every single URL in this way, but succeeded to identify a large majority. For example, URLs changed by link shorteners could not be processed.

As an example, for the URL ‘*https://dailystormer.ws/britain-to-give-houses-and-jobs-to-returning-isis-fighters/’* we would be able to use this approach to extract the words ‘*britain to give houses and jobs to returning isis fighters’*. This gives substantially more context for the remainder of the message that was posted alongside the URL and is likely to assist with hate speech detection.

## 1.3 Overview of Semantic and Syntactic features

Table SI 2 gives an overview of the semantic and syntactic features used in building the platform-specific hate speech models. Implementation of each feature is given in the second column.

*Table SI 2- Overview of semantic and syntactic features used in platform-specific models*

| | Feature | Implementation |
|---|---|---|
| **Semantic** | BERT Word Vectors (x 768) | rBERT (with Google Colab GPU) |
| | Plural Nouns / log odds | Apache OpenNLP / ‘tidylo’ R package |
| | Othering | Dictionary Approach |
| | Hate Terms | HateBase - Average scores |
| | Hate Severity | HateBase - Dictionary approach |
| | Hate Symbols | ADL - Dictionary approach |
| | Obscenity | Dictionary approach |
| | Sentiment | qdap implementation R |
| **Syntactic** | Lexical Density | Manual Calculation |
| | Flesch-Kincaid Reading Ease | Manual Calculation |
| | Document Length | Number of characters |
| | Punctuation | Count |
| | Negation | qdap implementation R |
| | Pronouns | Dictionary approach |

## 1.4 – Log-odds for plural nouns

We identified plural nouns in the messages using the Apache OpenNLP Maxent Part of Speech tagger from the ‘OpenNLP’ package (Hornik, 2016). To identify the relevance of these plural nouns for each label category, we calculated the weighted-log-odds of these nouns and included them as additional features in the model, using the ‘tidylo’ package (Schnoebelen, Silge, & Hayes, 2020). Log-odds are an effective way to represent words that capture differences in political speech and to evaluate the relative importance of those words (Monroe, Colaresi, & Quinn, 2008). They offer an advantage over the more traditional term frequency / inverse term frequency (tf-idf) for calculating the importance of words within a corpus as they do not penalise common words across all classes. This is beneficial in

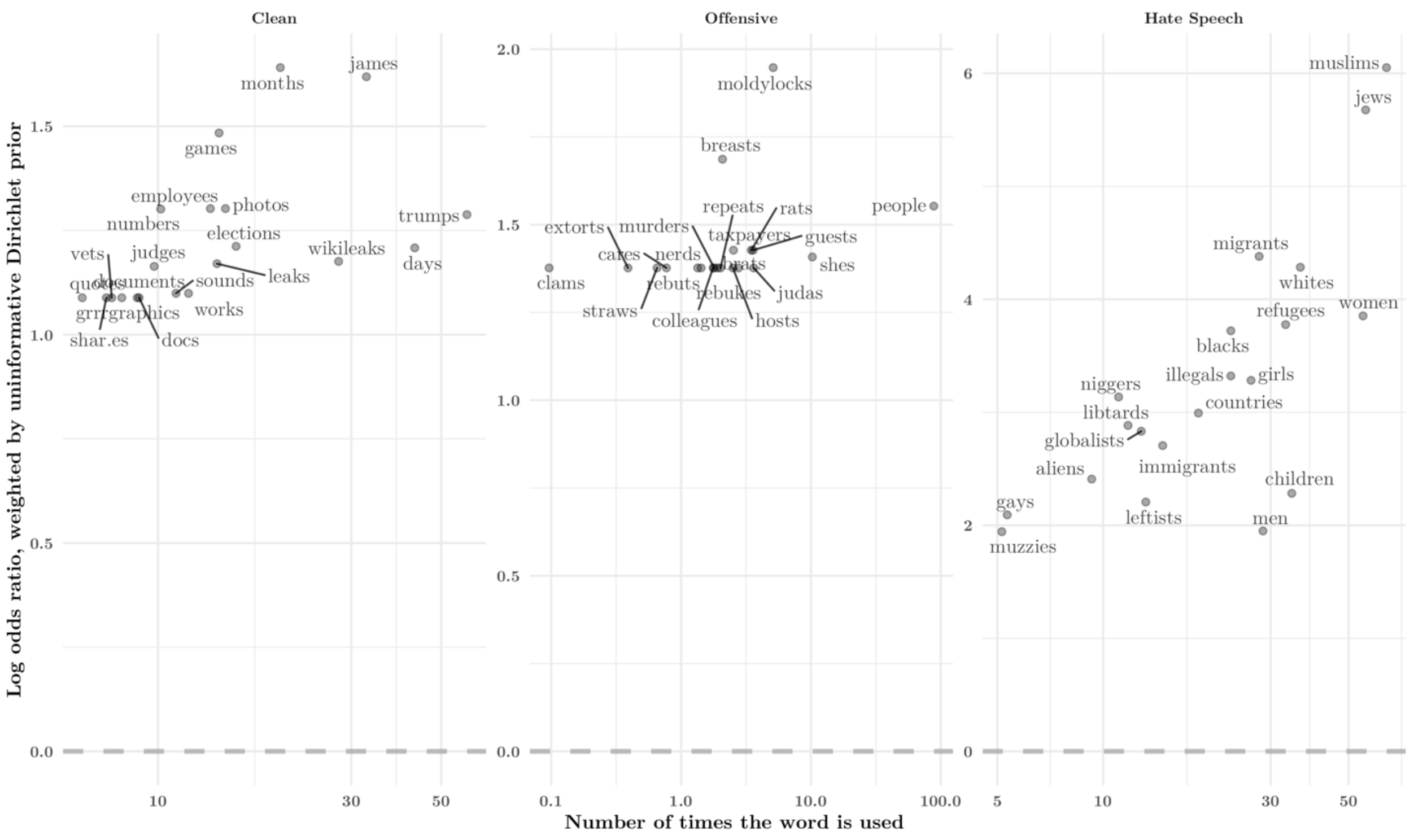

*SI Figure 1 - Log-odds values for plural nouns from across the three text classes within the Gab training data*

our case, as some plural nouns are common across the whole dataset but also informative when combined with other features and so should not be penalised (e.g. 'Muslims' will be common in both hateful and non-hateful discussions on the topic of religion).

SI Figure 1 shows the top plural nouns for each category within the Gab training data. Offensive and derogatory plural nouns, as well as standard terms for groups which often receive online abuse, receive high scores within the hate speech category, while the top terms in the clean category by contrast are not pejorative.

## 1.5 – Hyperparameter tuning and equivalence zones

In training all models, hyper-parameter tuning was done within the 10 fold cross validation, initially using a random grid search across standard values, and then tuned further during a second round focusing in on the values identified in the initial round (Kuhn & Johnson, 2013). For the xgboost platform-specific models, the final parameters for which a grid search was performed within the cross-validation are given below.

min_n – 10 : 30
mtry – 50 : 368 (Upper value equal to number of parameters after dimensionality reduction)
loss reduction – 10 : 30
learning rate - 0.05 : 0.1
number of trees – 1000 : 3000

Following model training, we defined an ordinal equivalence zone (Kuhn & Johnson, 2013) as when the difference in probability for the first (most likely) class and the second class was less than the inverse of the number of classes (in this case 0.33). Predictions in this zone were rejected as the models were not decisive enough.

## 1.6– Feature distinction for machine learning

SI Figure 2 shows the distribution of semantic and syntactic features across the three text categories (Clean, Offensive, Hate Speech) that were used in building the machine learning classifier. These values are normalised and so do not reflect the originally measured ranges. The high degree of overlap for all of these features suggests that in isolation they cannot be used and should be combined with broader semantic features.

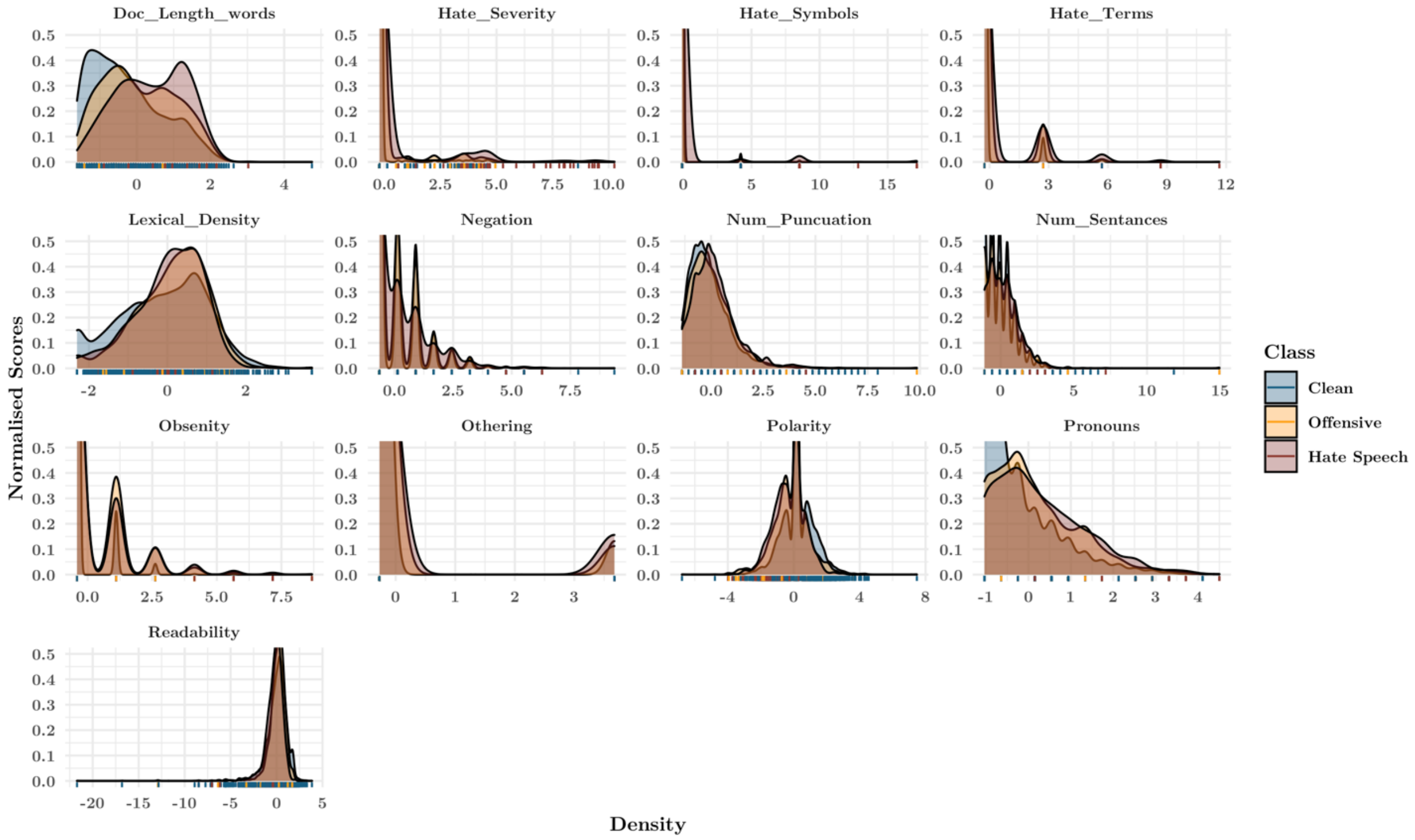

*Figure SI 2 – Normalised distributions of features used to train the platform-specific classifiers for the three message classes*

# 2 – Machine learning models performance

## 2.1 – Full platform-specific and superlearner model performance across three classes

Table SI 3 gives the performance metrics across all three classes for both the platform-specific models and the superlearner model. Accuracy, recall, precision and F1 scores are given for both the overall performance on the whole dataset, and specific performance on each social media platform.

## 2.2 ROC Curves for the Final Superlearner Model

Figure SI 3 shows the receiver operator characteristic curves (ROC curves) for the final superlearner model across the three classes of message.

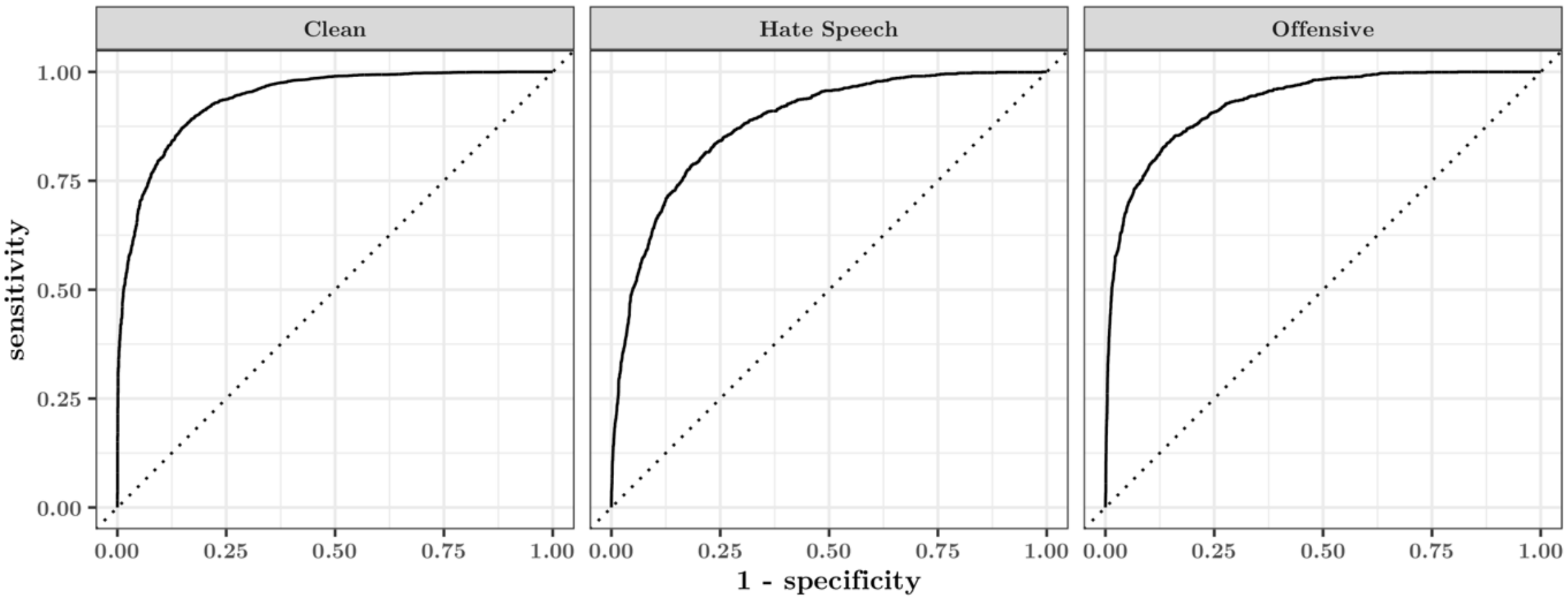

*Figure SI 3 – ROC Curves for the superlearner model across the three classes*

*Table SI 3 – Performance metrics (accuracy, precision recall and F1) for the three classes across platform specific and superlearner models*

| ***Platform*** | ***Model*** | ***Overall Accuracy*** | | ***Clean*** | | | ***Offensive*** | | | ***Hate Speech*** | | |
|---|---|---|---|---|---|---|---|---|---|---|---|---|
| | | **%** | **+/-** | **Precision** | **Recall** | **F1** | **Precision** | **Recall** | **F1** | **Precision** | **Recall** | **F1** |
| ***Overall*** | Platform Specific | 89.8 | 0.9 | 0.95 | 0.95 | 0.95 | 0.83 | 0.85 | 0.84 | 0.69 | 0.68 | 0.69 |
| | Superlearner | 91.4 | 0.8 | 0.95 | 0.97 | 0.96 | 0.86 | 0.86 | 0.86 | 0.73 | 0.66 | 0.69 |
| ***Facebook*** | Platform Specific | 89.3 | 1.8 | 0.94 | 0.95 | 0.94 | 0.79 | 0.64 | 0.71 | 0.75 | 0.83 | 0.78 |
| | Superlearner | 90.3 | 1.8 | 0.96 | 0.95 | 0.95 | 0.82 | 0.64 | 0.72 | 0.75 | 0.87 | 0.80 |
| ***Gab*** | Platform Specific | 89.8 | 2.1 | 0.95 | 0.96 | 0.95 | 0.66 | 0.58 | 0.61 | 0.73 | 0.75 | 0.74 |
| | Superlearner | 90.9 | 2.0 | 0.94 | 0.97 | 0.96 | 0.64 | 0.29 | 0.4 | 0.75 | 0.83 | 0.79 |
| ***Twitter*** | Platform Specific | 89.6 | 1.6 | 0.93 | 0.97 | 0.95 | 0.88 | 0.94 | 0.91 | 0.76 | 0.40 | 0.52 |
| | Superlearner | 89.9 | 1.6 | 0.94 | 0.95 | 0.95 | 0.88 | 0.96 | 0.92 | 0.71 | 0.36 | 0.48 |
| ***Stormfront*** | Platform Specific | 90.5 | 1.8 | 0.98 | 0.92 | 0.95 | 0.39 | 0.75 | 0.51 | 0.50 | 0.77 | 0.60 |
| | Superlearner | 93.9 | 1.3 | 0.96 | 0.98 | 0.97 | 0.67 | 0.58 | 0.62 | 0.67 | 0.42 | 0.52 |

## 2.3 Confusion Matrix for final Superlearner Model

Figure SI 4 shows the confusion matrix for the final super learner model across the three classes of message (Clean/Offensive/Hate Speech)

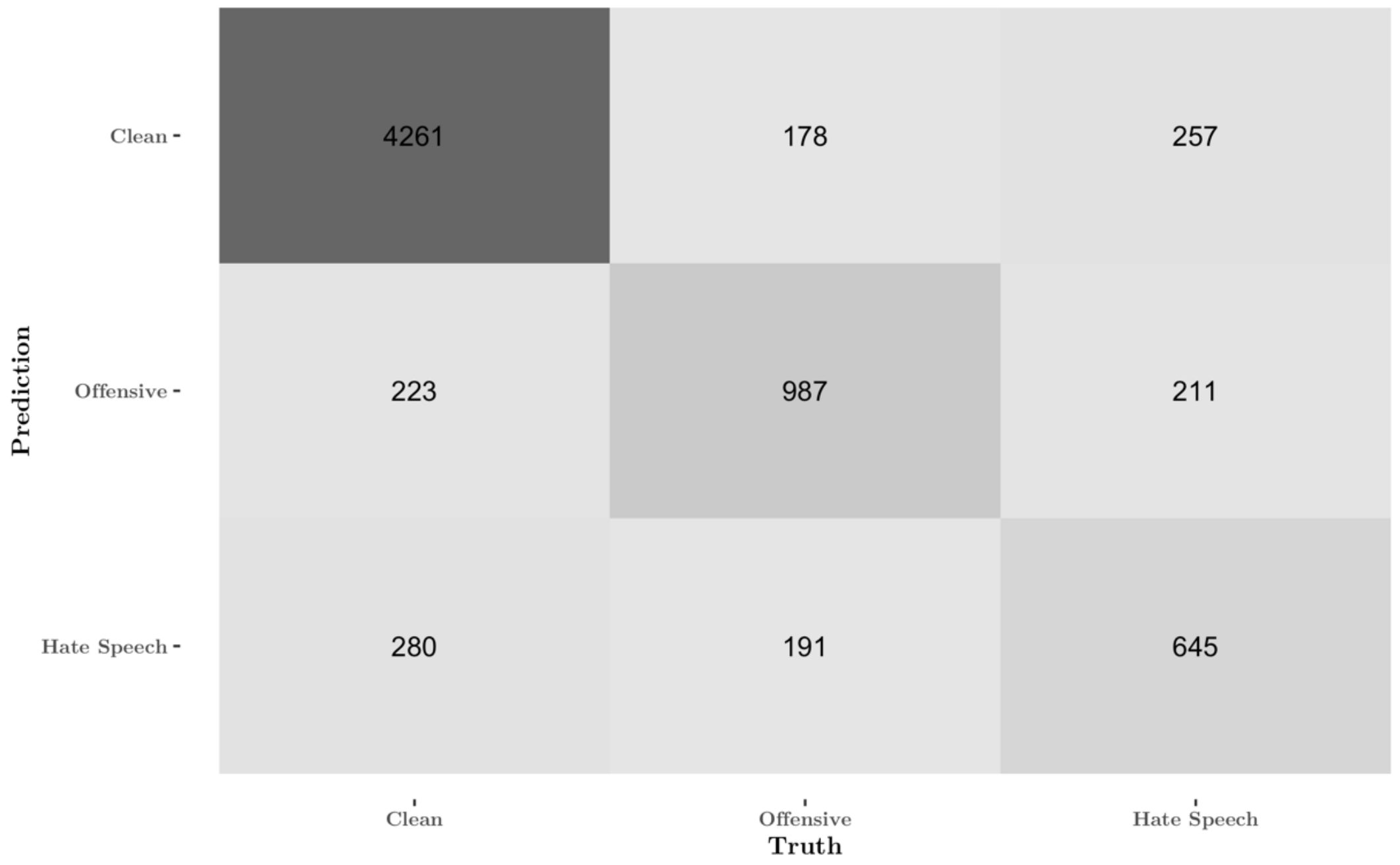

*Figure SI 4 – Confusion matrix for the superlearner model across the three classes*

## 2.4 Error inspection

When assessing accuracy values for the hate speech detection models, it is important to consider that not all incorrect classifications are equal, and minor category errors (i.e. between consecutive classes) is a smaller mistake than a major category error (between non-adjacent categories). To take this into account, we calculated the percentage of major category errors in the overall dataset at the superlearner level. We found that only 5.88% of clean messages were mislabelled as hate speech and 23.10% of hate speech messages were mislabelled as clean.

Investigating this further using the superlearner predictions on the Gab dataset as an example, removing the 'minor errors' increases performance in hate speech detection to 0.86 precision and 0.86 recall, up from 0.75 and 0.83.

## 2.5 Performance on novel Reddit data

Table SI 4 gives the performance metrics for the new Reddit data. These scores are given for the four types of model tested on this dataset: the HateSonar trained on Twitter data alone (Nakayama, 2018), a platform-specific model trained on Reddit data, the original superlearner presented in this paper (superlearner 1.0) and an updated superlearner model which includes the addition of the Reddit platform-specific predictions at the superlearner stage (superlearner 2.0).

***Table SI 4 – Performance metrics for Reddit data across four types of model; HateSonar, a platform-specific Reddit model, the original Superlearner and an updated Superlearner model with the inclusion of the Reddit platform specific model***

| ***Model*** | ***Overall Accuracy*** | | ***Clean*** | | | ***Offensive*** | | | ***Hate Speech*** | | |
|---|---|---|---|---|---|---|---|---|---|---|---|
| | % | +/- | **Precision** | **Recall** | **F1** | **Precision** | **Recall** | **F1** | **Precision** | **Recall** | **F1** |
| ***HateSonar*** | 61.6 | 3.1 | 0.87 | 0.84 | 0.85 | 0.89 | 0.37 | 0.52 | 0.19 | 0.81 | 0.30 |
| ***Reddit Platform Specific*** | 86.3 | 5.1 | 0.94 | 0.88 | 0.91 | 0.52 | 0.83 | 0.64 | 0.87 | 0.84 | 0.85 |
| ***Superlearner 1.0*** | 83.6 | 3.0 | 0.95 | 0.85 | 0.90 | 0.55 | 0.70 | 0.62 | 0.88 | 0.87 | 0.87 |
| ***Superlearner 2.0*** | 89.0 | 4.5 | 0.93 | 0.95 | 0.94 | 0.50 | 0.70 | 0.58 | 0.96 | 0.87 | 0.92 |

## 4 Cross platform (incongruent) performance on platform-specific models

Table SI 3 gives the performance for platform-specific models when giving prediction on incongruent platform data i.e. messages from the other three platforms which were not included in the training data. Predictions for the models on messages from that platform along with superlearner performance are also given in the main text in table 2.

Overall, we see that performance is lower when making predictions for data from other social media platforms. This suggests that the nature of the hate speech varies across platforms, both in semantic and syntactic features. We did not observe a drop in performance for models on test sets compared to cross-validation so this is not due solely to over-fitting to the training data although we cannot rule this out completely or that the overfitting does not occur in the sampling approach (i.e. over sampling a certain type of hate speech from a specific platform).

Despite this overall reduction, we also observe substantial variation depending on which pair of platforms are being compared. We find for example that the performance is reasonably high when comparing Gab and Facebook models in both directions suggesting that these platforms share in the nature of the hate speech that they host, while predicting Twitter hate speech from Stormfront models performs poorly as does predicting Gab hate speech from the Twitter model. The latter of these gives particularly low levels of hate speech recall with nearly 9 out of every 10 hate speech messages missed. Indeed, we find that the first of these (Twitter models predicting Stormfront messages) performed at chance levels. All other models performed above chance overall however, showing that there is still useful information shared across platforms and why the superlearner approach allows us to improve performance overt single platform models.

Investigating this in more detail we find that the distribution of hate speech predictions for each classifier varies depending on the platform that hate speech messages come from. Figure SI 5 shows this for predictions across all 5 platforms (this time with the addition of the Reddit data) for the Gab specific model. This shows that platforms in the top row (Facebook, Reddit and Gab) have similar levels of distinction in predictions between classes, and therefore information is being used to make meaningful distinctions, while platforms in the bottom row (Stormfront and Twitter) have much less clear distinctions in the prediction distributions across classes.

*Table SI 4- Comparison of model performance in detecting hate speech for datasets from different social media platforms*

| Model \ Data | Facebook | Gab | Twitter | Stormfront |
|---|---|---|---|---|
| **Facebook** | Accuracy – 82.3% | 85.6 | 72.3 | 93.1 |
| | HS Precision – 74.5 | 57.9 | 31.3 | 63.2 |
| | HS Recall – 82.7 | 57.9 | 12.6 | 38.7 |
| **Gab** | 86.7 | 89.8 | 74.3 | 91.6 |
| | 67.6 | 73.0 | 34.0 | 52.8 |
| | 75.8 | 75.0 | 14.1 | 68.8 |
| **Twitter** | 81.3 | 78.5 | 89.6 | 90.1 |
| | 66.2 | 66.7 | 76.2 | 62.5 |
| | 28.7 | 11.0 | 39.8 | 26.9 |
| **Stormfront** | 88.1 | 84.4 | 73.5 | 90.5 |
| | 64.8 | 58.8 | 29.2 | 49.6 |
| | 76.7 | 62.2 | 14.7 | 76.7 |

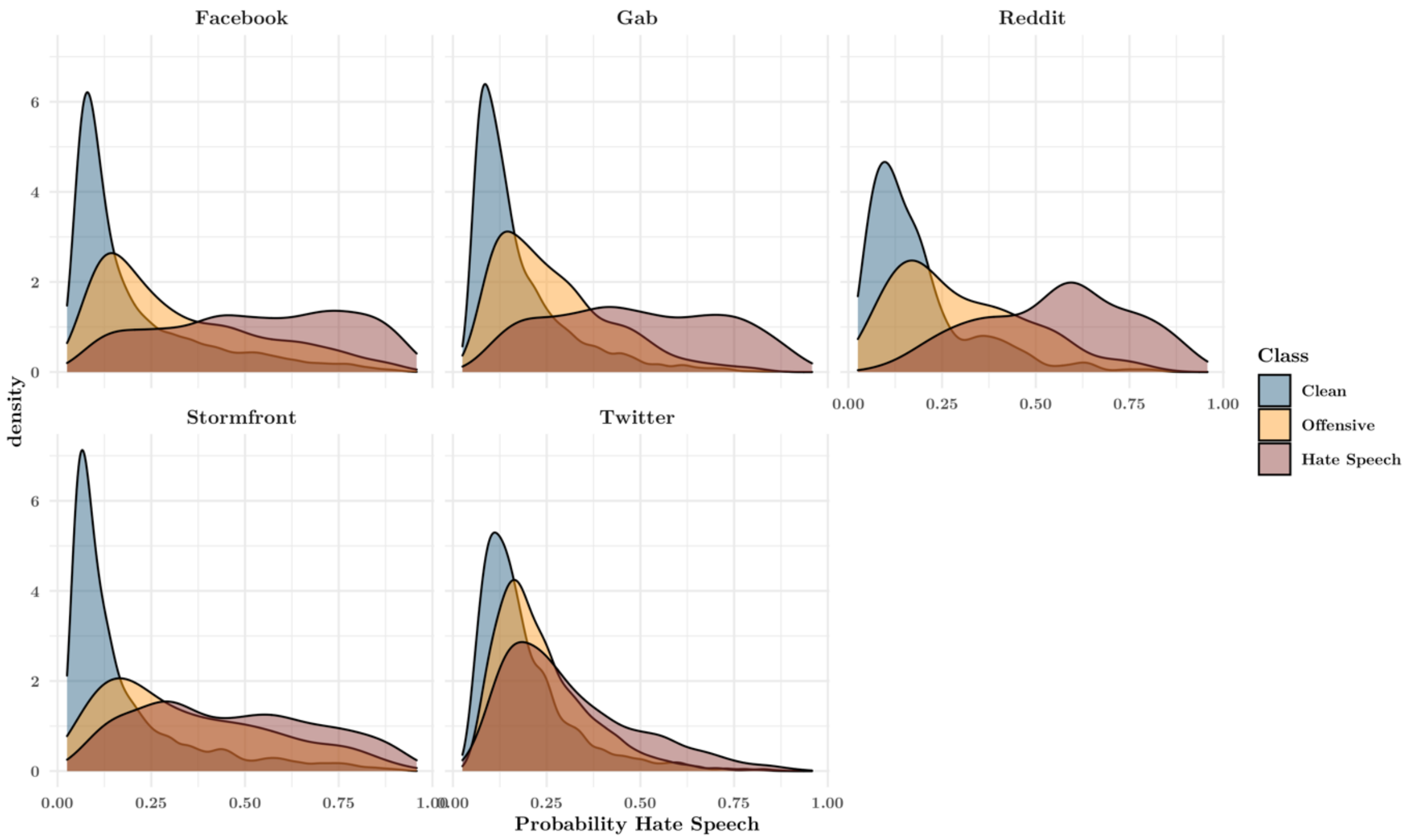

*SI Figure 5 – Distribution of hate speech predictions for the Gab specific classifier for the five platforms across the three messages classes*

## SI References